\pdfoutput=1

\documentclass[11pt]{article}
\usepackage{ACL2023}

\usepackage{times}
\usepackage{latexsym}
\usepackage{booktabs}
\usepackage[T1]{fontenc}
\usepackage[utf8]{inputenc}
\usepackage{microtype}
\usepackage{inconsolata}
\usepackage{soul}
\usepackage{cleveref}
\usepackage{xspace}
\usepackage{enumitem}
\usepackage{algorithm}
\usepackage{algpseudocode}
\usepackage{siunitx}
\usepackage[disable]{todonotes}

\newcommand{\ie}{i.e.,~}
\newcommand{\eg}{e.g.,~}

\newcommand{\example}[1]{{``\emph{#1}''}}
\newcommand{\exampleb}[1]{{\textcolor{blue}{``\emph{#1}''}}}

\newcommand{\OURMODEL}{\textsc{Surf}\xspace}
\newcommand{\REP}{\textsc{Rep}\xspace}
\newcommand{\ROO}{\textsc{Roo}\xspace}
\newcommand{\GEN}{\textsc{Gen}\xspace}
\newcommand{\ROOGEN}{\textsc{Roo+Gen}\xspace}
\newcommand{\ORACLE}{\textsc{Optimal}\xspace}

\title{Answering Unanswered Questions through Semantic Reformulations\\ in Spoken QA}

\author{
Pedro Faustini$^1$\thanks{~~Work done during an internship at Amazon.}, Zhiyu Chen$^2$, Besnik Fetahu$^2$, Oleg Rokhlenko$^2$,  Shervin Malmasi$^2$ \\
 $^1$Macquarie University, Sydney, NSW, Australia\\
 $^2$Amazon.com, Inc., Seattle, WA, USA\\
\resizebox{\linewidth}{!}{
    \texttt{pedro.arrudafaustini@hdr.mq.edu.au} ~~~~ \texttt{\{zhiyuche,besnikf,olegro,malmasi\}@amazon.com}
    }\\
}

\begin{document}

\maketitle

\begin{abstract}
Spoken Question Answering (QA) is a key feature of voice assistants, usually backed by multiple QA systems.
Users ask questions via spontaneous speech which can contain
disfluencies, errors, and informal syntax or phrasing.
This is a major challenge in QA, causing unanswered questions or irrelevant answers, and leading to bad user experiences.
We analyze failed QA requests to identify core challenges:
lexical gaps, proposition types, complex syntactic structure, and high specificity.
We propose a Semantic Question Reformulation (\OURMODEL) model offering three linguistically-grounded operations (repair, syntactic reshaping, generalization) to rewrite questions to facilitate answering.
Offline evaluation on 1M unanswered questions from a leading voice assistant shows that \OURMODEL significantly improves answer rates:
up to 24\% of previously unanswered questions obtain relevant answers (75\%).
Live deployment shows positive impact for millions of customers with unanswered questions;
explicit relevance feedback shows high user satisfaction.

\end{abstract}

\section{Introduction}

Question Answering (QA) is a longstanding NLP task, and voice assistants like Alexa have made Spoken QA ubiquitous.
Users often address such assistants with spontaneous speech, as they would a human.
However, differences between spoken and written language \cite{chafe1987relation}, such as the presence of disfluencies, informal or incomplete speech, and different syntax have been shown to pose challenges for NLP tasks \cite{ward-1989-understanding,shriberg2005spontaneous,salesky2019fluent}.
QA system mostly use written data, and such phenomena impact question understanding and answer retrieval \cite{gupta2021disfl}, leading to irrelevant answers or unanswered questions, leaving users unsatisfied.

Recently, language generation has been used to improve QA through Question Rewriting (QR).
For example, QR is used in conversational systems to answer contextual questions in multi-turn dialogues~\cite{ye2022robustness}.
While QA models can be improved with fine-tuning, real-world systems have multiple QA backends and retraining is expensive, making input rewriting a practical solution~\cite{chen-etal-2022-reinforced}. This has the added benefit that a single QR model may improve multiple QA systems.

\begin{figure}
    \centering
    \includegraphics[width=0.99\columnwidth]{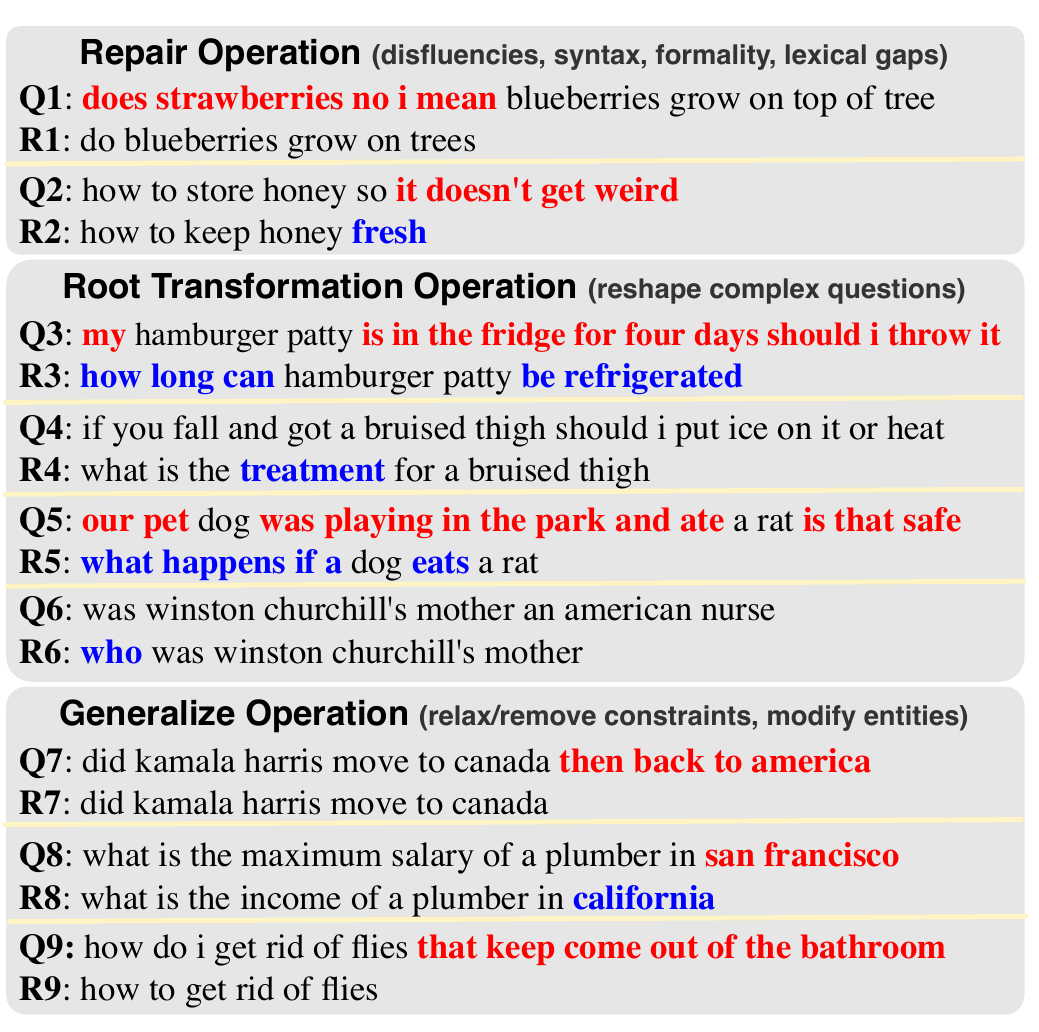}\vspace{-10pt}
    \caption{\textmd{Examples of challenging questions (Q) and our proposed reformulation operations (R) on them.}} 
    \label{fig:reformulation_example}
\end{figure}

We propose applying QR to reformulate difficult or unanswered questions. 
We analyzed millions of answered and unanswered real-world questions from a leading voice assistant to understand the factors impacting QA failure (\S\ref{sec:challenges}).
In addition to the well-known issue of disfluencies, we identify novel challenges from question structure and specificity.
To address them, we propose three linguistically-informed reformulation operations that only require the question (\S\ref{sec:method}).
The operations, shown in \Cref{fig:reformulation_example}, are designed to \emph{improve answerability}\footnote{We consider a question answerable if a QA system decides with sufficient confidence that a suitable answer is retrieved.} based on common speech patterns, so that for a previously unanswered question, the same QA system is able to provide an answer for its reformulation.

While question repair has been studied, our \textit{root wh-} and \textit{question generalization} operators are novel contributions of this work.
Our results demonstrate that our approach can achieve:

\begin{enumerate}[leftmargin=*]
    \itemsep0em
    \item high reformulation accuracy of 83\% for rewriting questions to a desired shape (\S\ref{subsec:intrinsic_evaluation});
    \item improving the answer rate of previously unanswered questions by up to 24\% (\S\ref{subsec:question_answerability}); and
    \item 75\% of answers on reformulated questions are relevant to the original question (\S\ref{subsec:relevance}).
\end{enumerate}

Live deployment of our model (\S\ref{sec:deployment}) achieves positive impact for millions of users with unanswered questions, and explicit relevance feedback from customers shows high satisfaction.
\section{Related Work}
\label{sec:related_work}

\vspace{-4pt}
\paragraph{Question Quality:} 
QA models are typically trained on formal written language, and are known to be impacted by the quality of user questions.
An analysis of the WikiAnswer dataset \cite{Fader14} by \citet{10.1145/3357384.3358046} showed that 68\% of the questions were ill-formed, usually due to wrong words, wrong order, or background noise, harming the answerability of those questions.
\citet{gupta2021disfl} examined the impact of disfluencies in QA, showing that they had a large impact on answering performance.
Many of these issues stem from natural properties of spontaneous speech, such as errors, self corrections, and informal syntax \cite{chafe1987relation}.
Our work tackles these issues, and tries to go beyond corrections by considering question types and question specificity.

\vspace{-4pt}
\paragraph{Question Complexity:} Depending on the QA system, some questions may be more difficult to answer. It has been shown that questions requiring multi-hop reasoning are more challenging~\cite{yang-etal-2018-hotpotqa}, often leading to no answers or wrong answers.
Questions are affected by the broader types of syntactic complexity explored in the field \cite{nassar-etal-2019-neural, martin-etal-2020-controllable, sheang-saggion-2021-controllable}.
Regardless of complexity, questions may also be unanswerable due to incorrect framing or false suppositions~\cite{kim-etal-2021-linguist}.
Other work has analyzed questions in different datasets, showing that \emph{wh-*} words (e.g. \emph{who, what, when}) are the dominant way to start a question \cite{ko-etal-2020-inquisitive},
and that these words and related phrases (\eg \example{how much}, \example{how large}) are associated with reduced answering complexity \cite{chali-hasan-2012-simple}.
In our work we consider how controlled syntactic restructuring can address the above challenges to reduce answering complexity.

\vspace{-5pt}
\paragraph{Question Rewriting:}
Rewriting questions is a natural extension of query reformulation approaches used to improve Information Retrieval \cite{he2016learning}.
Question rewriting has been applied to improve QA in different ways.
Question paraphrasing has been used as a data augmentation approach to retrain QA systems to improve robustness \cite{gan-ng-2019-improving}.
\citet{buck2018ask} propose using a reinforcement learning agent between the original question and a black box QA system. The agent probes the QA system with several reformulations to learn how to elicit the best answer.
\citet{10.1145/3357384.3358046} propose a question refinement system to rewrite malformed questions.

\vspace{-5pt}
\paragraph{Rewriting Operations:}
Text rewriting is based on specific linguistic changes.
\citet{nassar-etal-2019-neural} note that text simpliﬁcation changes can be lexical (rare words replaced by more common ones) and syntactic (complex structures are split, reordered, or deleted).
\citet{tomuro-2003-interrogative} notes that paraphrasing questions is more difficult as the interrogative structure is separate from the declarative, and can have many variations. They quantified paraphrasing operations and showed that interrogative reformation accounted for 50\% of changes, followed by lexical substitution (25\%) and semantic changes (16\%).
Recent work on sentence rewriting has followed this direction, by breaking down reformulation into predefined editing operations \cite{choi-etal-2021-decontextualization}.

Our work is inspired by all of the above, but differs in several ways.
We expand on the known issues in QA by analyzing real voice assistant data to identify prevalent challenges to tackle; we consider malformed question correction as a prerequisite for dealing with challenges of complex questions. 
Additionally, prior rewriting approaches aim to improve QA via retraining, or by building a rewriter tailored to a single QA system.
We take a different approach that does not rely on answer data or QA system feedback, and build a general model that can benefit multiple QA systems in a federated architecture.
Instead of uncontrolled paraphrasing, we deal with question complexity via controllable reformulations that distinguish between lexical modification,
interrogative clause restructuring, and semantic changes.
We propose novel linguistic restructuring operations to deal with complex syntax, and generalize high-specificity elements. 

\section{Challenges in Real-world Spoken QA}
\label{sec:challenges}

First, to quantify and understand why spoken QA fails, we perform a failure analysis on 10 million real questions, by further distinguishing questions according to their \emph{question type} (we define 5 types based on linguistic properties, see \Cref{sec:qtype_heuristic} for details) from a leading voice assistant.

\textbf{Scope:} we limit our work to questions that were not answered due to retrieval failure, but may potentially have relevant answers if reformulated.
They must be valid questions (seek knowable knowledge) whose information need can be understood (by humans) and re-stated.
QA may fail for other reasons; we do not consider such issues \eg \textit{inter alia}, ASR errors, invalid or difficult to understand questions, subjectivity, and other reasons for retrieval failure.%

A quantitative and qualitative study was undertaken by domain experts (details in \Cref{sec:failure_analysis}), and identified the below challenges (C1-7) as contributing to a significant proportion\footnote{The exact numbers cannot be divulged for confidentiality.} of failed requests, and potentially solvable by reformulation.
\vspace{3pt}

\noindent\textbf{C1. Malformed Utterances:}
Questions with disfluencies and syntactic errors were more likely to fail \eg Fig.~\ref{fig:reformulation_example} (Q1).
Correction methods have previously been used to fix these \cite{gupta2021disfl}.

\vspace{3pt}
\noindent\textbf{C2. Lexical Gaps}
Questions framed colloquially or lacking appropriate parlance for a topic \eg Fig.~\ref{fig:reformulation_example} (Q2), were associated with failure.
This is caused by lexical gaps \cite{riezler-liu-2010-query} arising from language mismatch between the user input and answer sources, as QA systems use formal knowledge sources for retrieval. Lexical substitution and rephrasing may address this challenge.

\vspace{3pt}
\noindent\textbf{C3. Complex Syntactic Structure:} Utterances with complex structure, such as multi-clause questions, can lead to QA failure.
Such phrasing is more common in spoken language, and can be simplified via syntactic restructuring, \eg Fig.~\ref{fig:reformulation_example} (Q3-5).

\vspace{3pt}
\noindent\textbf{C4. Polar Propositions:} Yes-No questions are asked to confirm a specific proposition, \eg \example{Do box turtles live in Japan?}.
Answering polar questions is more difficult than wh-questions for both humans \cite{moradlou2021wh} and QA systems \cite{DBLP:conf/naacl/ClarkLCK0T19}, due to the entailment and inferences required to arrive at an answer.
This can be simplified by reformulating to a factoid wh-question, \eg \example{Where do box turtles live?}.

\vspace{3pt}
\noindent\textbf{C5. False Presuppositions:} Additionally, polar questions may contain false presuppositions that cause retrieval failure \cite{kim-etal-2021-linguist}.
We hypothesize rewriting such questions to wh-questions may retrieve relevant answers, \eg Fig.~\ref{fig:reformulation_example} (Q6).

\vspace{3pt}
\noindent\textbf{C6. High Specificity:}
Highly specific questions (concerning very specific entities, or conditions) may not be answerable.
We believe generalizing such questions by entity modification or constraint relaxation (Fig.~\ref{fig:reformulation_example} Q7-9) can broaden answer recall.

\vspace{3pt}
\noindent\textbf{C7. Irrelevant Info:} Related to C3 and C6, complex and high-specificity questions may contain contextual facts that are irrelevant to the answer. We believe removing such details can improve answer recall (Fig.~\ref{fig:reformulation_example} Q4/5/7).
\section{\OURMODEL Question Reformulation Model}
\label{sec:method}

We now describe our proposed \textbf{S}emantic Q\textbf{u}estion \textbf{R}e\textbf{f}ormulation model (\OURMODEL) and the reformulation operators that it supports.

\subsection{Reformulation Model}
\label{subsec:query_reformulation_model}

Inspired by controllable multi-task learning for text generation \cite{keskar2019ctrl,2020t5}, we train a single model to perform different reformulations.
Our reformulation model, $\mathcal{F}(p,q)$, represents a seq2seq Transformer model~\cite{lewis-etal-2020-bart}, and is trained such that for an \emph{input} question $q$ and a \emph{target} reformulation operator $p\in\{$\REP, \ROO, \textsc{Gen}$\}$, pre-pended as a prefix to $q$, it reformulates $q$ into $q'$ according to $p$.

\noindent\textbf{Model Training.} $\mathcal{F}$ is trained in two stages: the first stage pretrains $\mathcal{F}$ using a large \emph{weakly-supervised} corpus (derived by a heuristic proposed in \S\ref{sec:data}) of  $\langle q, q'\rangle$ for the \REP  and \ROO operations.
In the second stage, we finetune $\mathcal{F}$ on manually annotated pairs of $\langle q, q'\rangle$ for all operators in $p$.

\subsection{Reformulation Operators}
\label{subsec:query_reformulation_operators}

Each prefix $p$ instructs $\mathcal{F}$ to perform a specific type of reformulation. We define the following prefix operators based on the challenges presented in \S\ref{sec:challenges}.

\vspace{3pt}
\noindent\textbf{Question Repair (\REP):}
To address challenges \textbf{C1-2}, \REP removes disfluencies, performs syntactic correction, and increases formality via lexical substitution with high-entropy words.
For example, the input \example{Where can I get a booze after 11 pm?} is repaired to \example{Which stores sell beer after 11 pm?}.

\pagebreak

\noindent\textbf{Root Wh- Transform (\ROO):} 
Outlined in challenges \textbf{C3-6}, questions with complex structure are more difficult, but may be answered if simplified to factoid questions.
\ROO reformulates $q$ such that the interrogative \emph{wh-*} phrase is clause initial (at the root of the sentence), making needed syntactic adjustments. For example, 
\example{Do any universities in Germany offer degree programs taught in English?}
$\rightarrow$ \example{Which universities in Germany offer degree programs in English?}.
This also handles contextualized multi-clause questions, \eg \example{I am chopping onions for a pizza dinner how fine should they be} $\rightarrow$ \example{how fine should onions be for pizza}.

\vspace{3pt}
\noindent\textbf{Question Generalization (\GEN):}
To deal with highly specific questions, covered in challenge \textbf{C4}, we propose a novel question generalization operation.
Inspired by similar approaches to improve recall in structured query languages \cite{motro1984query} and IR \cite{boldi2011query}, we simplify questions through the removal or relaxation of semantic constraints. Creating a more general question allows the retrieval of a superset of results, which in many cases provides a highly related answer that may be better than no answer.
\GEN does this by dropping adjuncts, replacing nouns with hypernyms or holonyms, and removing adjectives.
For example, the question \example{Do poisonous pythons live in Miami?} can be generalized to \example{Do snakes live in Florida?}. Note that \example{python} and \example{Miami} are turned into more generic entities \example{snake} and \example{Florida}, and at the same time the aspect \example{poisonous} is dropped.

In \ROO and \GEN, \REP is always performed jointly with the respective operators. The output of all operators should not contain any syntactic or semantic errors present in the original question.

\section{Experimental Setup}

\subsection{Intrinsic Evaluation Strategy}

Using a human study, we intrinsically evaluate the \emph{reformulation accuracy}\footnote{Note that due to the unreliability of automated metrics such as BLEU and ROUGE, we opt for human evaluation and answering metrics in our intrinsic and extrinsic evaluations.} to assess if: (1) the reformulation \emph{retains} the intent of the input question; and (2) the reformulation \emph{satisfies} the properties of the reformulation operator $p$.  

\paragraph{Evaluation Data:} For each question type, we randomly sampled $50$ questions from each reformulation operator.
This data is then assessed by expert annotators, resulting in $1,000$ annotations.

\subsection{Extrinsic Evaluation Strategy}
For the extrinsic evaluation, we assess the impact of the reformulated questions on two aspects:

\vspace{-5pt}
\paragraph{\textbf{Answer Rate:}} measured as the percentage of reformulations that obtain an answer.

\vspace{-5pt}
\paragraph{\textbf{Answer Relevance:}} a three-point scale measuring the answer relevance to the original question (obtained from the reformulated questions): \textbf{Irrelevant} (0): answer is not related to $q$;  \textbf{Related} (1): answer is partially relevant;\footnote{\eg it answers the question's intent but the subjects may be different, as may be the case of entities in \GEN.} and  \textbf{Exact} (2): answer exactly satisfies question's information need.

\vspace{-5pt}
\paragraph{Evaluation Data:} For the two aspects we measure, we consider the following evaluation datasets:

\vspace{-7pt}
\begin{itemize}[leftmargin=*]
    \itemsep0em
    \item \textbf{Answer Rate:} We randomly sampled 1M \emph{unanswered} questions by our QA system (see \Cref{apdx:eval} for additional details).
    \item \textbf{Answer Relevance:} on the same questions used for intrinsic evaluation, the annotators also check the answer relevance w.r.t the original question.
\end{itemize}

\subsection{Training Data}
\label{sec:data}

\noindent\textbf{Pre-training Data.} We create a weakly-supervised dataset of $1.2$M samples, derived from the MQR corpus \cite{chu-mqr-20}, which provides tuples of ill-formed and well-formed questions (c.f. \S\ref{sec:pretrain}). To construct input tuples $\langle p, q\rangle$ for pre-training $\mathcal{F}$, from a target question $q'$ we derive $p$ as follows. First, using the algorithm in \Cref{sec:qtype_heuristic}, we identify the question types of $q$ and $q'$.  If $q$ and $q'$ have the same type, then $p=$ \REP. If $q'$ is a \emph{root} question and $q$ is not, then $p=$ \ROO. 
The \GEN operator is novel to our work and cannot be automatically derived, and is part of the fine-tuning dataset.

\noindent\textbf{Fine-Tuning Data.} We sampled $3,851$ questions and annotated reformulations, based on guidelines listed in \S\ref{sec:sop}, for all operators in \S\ref{subsec:query_reformulation_operators}.
We use 10\% of annotated data for validation; the rest is used during the second stage of training to fine-tune $\mathcal{F}$.

\subsection{Reformulation Model Configurations}

\noindent\textbf{\OURMODEL:} At inference time, our model\footnote{\Cref{subsec:implementation_details} shows training and hyperparameter details.} can do different reformulations based on $p$. We analyze the impact on question answerability from different reformulation operators \REP, \ROO and \GEN. Additionally,  we analyze the combination of \ROO and \GEN, \ie $q$ is first reformulated by \ROO, then the resulting $q'$ is reformulated by \GEN, denoted as \ROOGEN. Note that the operators \ROO, \GEN, \ROOGEN, all perform a \REP operation as well (see Section \ref{subsec:query_reformulation_operators} for details).

\noindent\textbf{Baseline:} As a baseline model we consider an ablation of \OURMODEL without its pre-training stage, assessing its performance on the same four operators. 

\noindent\textbf{\ORACLE:} We consider the case where $q$ is answered if \emph{any} of its reformulations $p\in\{$\REP, \ROO, \textsc{Gen}$\}$ obtains an answer. \ORACLE represents the upper bound performance of the QA system.\footnote{Live deployment latency requirements prohibit  producing all possible reformulations and running through a QA system; therefore we try to determine the best single operator $p$ offline.}

\section{Results and Discussion}
\label{sec:evaluation}
We now turn to a discussion of the results for the \emph{intrinsic} (accuracy) and \emph{extrinsic} (answer rate and relevance) evaluation strategies.

\subsection{Intrinsic Evaluation}\label{subsec:intrinsic_evaluation}

\begin{table}[ht!]
\centering
\resizebox{1\columnwidth}{!}{
\begin{tabular}{l | c |  lll}
\toprule
\textbf{Task} & \textbf{Accuracy} & \multicolumn{3}{c}{\textbf{Answer  Relevance}}\\ 
\midrule
& &  Irrelevant & Related & Exact \\

\midrule
\ROOGEN & $77\%$  & 26\% (4.6\%) & 25\% \textbf{(4.5\%)} & 48\% \textbf{(8.6\%)}  \\
\GEN & $\mathbf{83\%}$ & 29\% (4.2\%) & 22\% (3.2\%) & 49\% (7.0\%)\\
\ROO & $73\%$ & 36\% (4.7\%) & 18\% (2.4\%) & 46\% (6.1\%)\\
\REP & $80\%$ & 26\% (2.5\%) & 15\% (1.4\%) & 59\% (5.5\%) \\ 
\bottomrule
\end{tabular}}
\caption{Evaluation results from the human study on \emph{reformulation accuracy} and \emph{answer relevance}. For answer relevance, in brackets are shown the \emph{extrapolated estimations} of the absolute percentages of answered questions from Table~\ref{tab:question_types} and their respective answer relevance. \ROOGEN obtains the highest answer rate and relevance with 13.1\% or 131k questions.
}\label{t:human}
\end{table}

\Cref{t:human}
shows the human evaluation results for reformulation accuracy.
The best accuracy is achieved for \GEN, with 83\% of the reformulations being accurate. 
This is because \GEN does not require changing the question type like \ROO.

\REP achieves second best accuracy. One reason for the slightly lower accuracy than \GEN, is that it sometimes changes the question type (e.g. request to root), which goes beyond the \REP's reformulation scope. Although according to our intrinsic evaluation strategy such cases represent inaccurate reformulation, in practice this is benign as QA systems perform very well on root factoid questions.

Finally, we note that reformulations significantly shorten the input questions and result in higher type-token ratio (\Cref{sec:change_analysis}).
We list many examples of model input/output pairs in \Cref{sec:examples}.

\begin{table}[t]
\centering
\resizebox{0.65\columnwidth}{!}{
\begin{tabular}{l r r}
\toprule
\textbf{Task} & \textbf{Baseline} & {\textbf{\OURMODEL}} \\
\midrule
No reformulation     & 0.00\%     & 0.00\% \\
\midrule
\REP        & 8.10\%  & 9.41\%\\
\ROO        & 9.26\%  & 13.18\%\\
\GEN        & 13.29\% & 14.34\%\\
\ROOGEN     & 14.50\% & 17.92\%\\
\midrule%
\ORACLE     & 18.48\%  & 24.15\%\\
\bottomrule
\end{tabular}}
\caption{Results for the baseline and \OURMODEL models using different reformulation types on our test set.}\label{rs:main}
\end{table}

\subsection{Extrinsic Answer Rate Results}
\label{subsec:question_answerability}

\begin{figure}[t!] %
\includegraphics[width=\columnwidth]{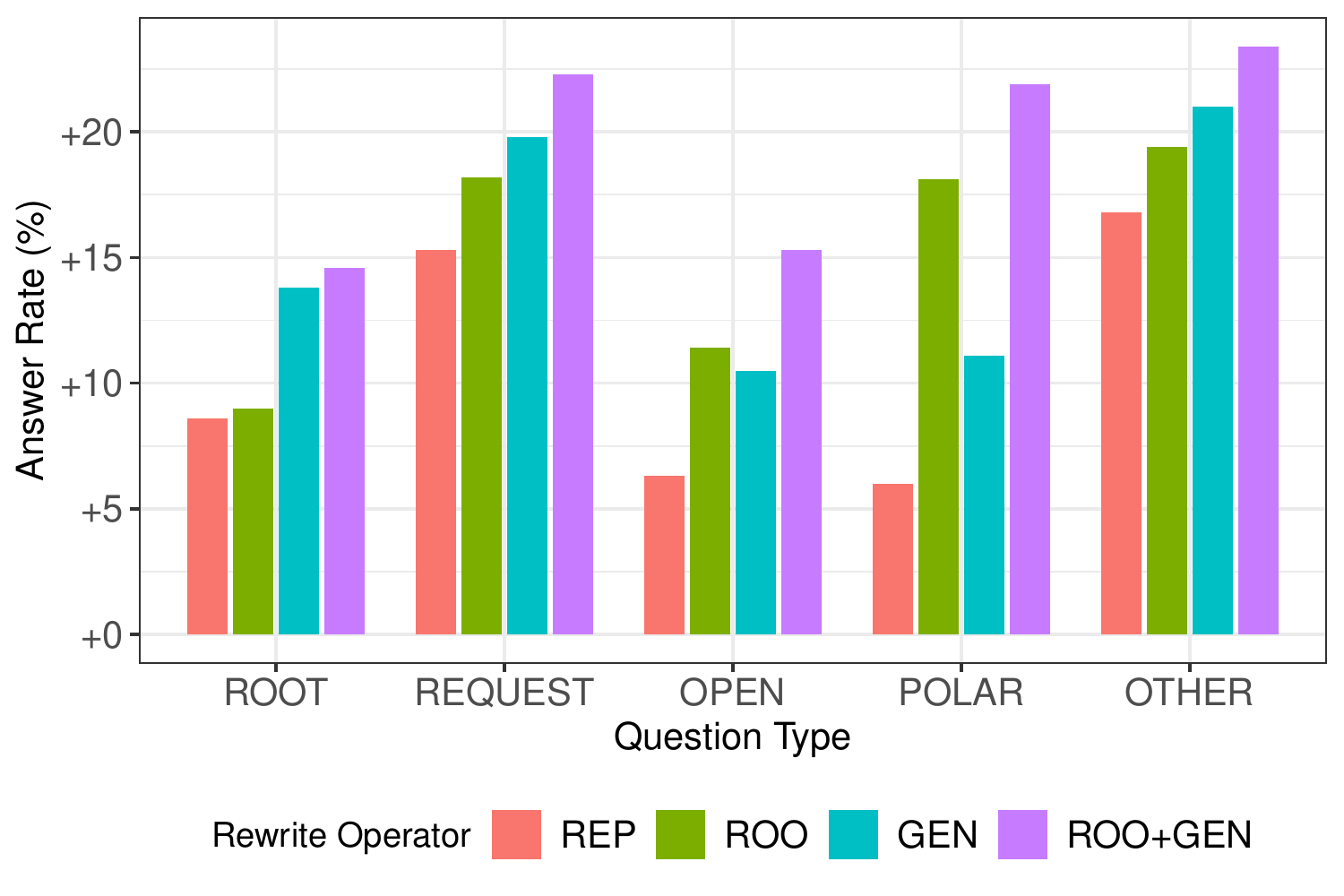}
\caption{Answer rate of different reformulation tasks grouped by original question types.}
\label{fig:answer_rate}
\end{figure}

Table~\ref{rs:main} shows results for all reformulation tasks and models.
\ORACLE represents the case where for an input question at least one reformulation operator gets answered by the QA system.
Pre-training yields consistent improvement in all tasks. Our large weakly supervised $\langle q, q'\rangle$ data enables learning the \REP and \ROO operations, leading to an answer rate improvement for \OURMODEL-\ROO with 13.18\% over the Baseline of 9.26\% (a 3.92\% absolute improvement).
\Cref{fig:answer_rate} shows a breakdown of the impact of the operators by question type.

\paragraph{Impact of Speech Errors:}
the \REP operation, which performs correction and makes question more formal, shows a consistent answer rate improvement across all tasks and models, improving it by 9.4\%.
This demonstrated that for many questions speech errors and framing cause retrieval failure. In \Cref{fig:answer_rate} we note that \REP provides a consistent improvement across all question types.
This improvement is intuitive given that a core component of QA systems is their ability to understand questions before answering, hence any speech or syntactic errors negatively impact answering.

\paragraph{Impact of Root Transformation:}
the \ROO operation repairs and reformulates the question to its root form. It shows better performance than \REP, although it may change the original question type. For \OURMODEL, the improvement of \ROO over \REP are with 3.77\%, contrary for baseline where the improvement is only 1.16\%. This further highlights the importance of the pre-training stage for \OURMODEL.
Figure~\ref{fig:answer_rate} shows that for all question types, reframing them as {root} questions significantly improves the answer rate.
\ROO is the most effective operator for polar questions, as they are particularly are hard to answer (\S\ref{sec:challenges}).
For example, \example{Is Sherlock Holmes a real person?} can also be answered via the alternative question \example{Who is Sherlock Holmes?}.

\vspace{-5pt}
\paragraph{Impact of Generalization:} the \GEN operation repairs and generalizes the original question to be less specific. For \OURMODEL, \GEN obtains 4.93\% absolute improvement over \REP in terms of answer rate, similar is improvement for the baseline with 5.19\% (cf. Table~\ref{rs:main}).
As we show in \S\ref{subsec:relevance}, most of the provided answers to the generalized questions are in fact relevant to the original question's intent.

\vspace{-5pt}
\paragraph{Impact of Joint Reshaping and Generalization:} 
\ROOGEN achieves the best performance across all tasks. This is intuitive as questions are first corrected for possible errors, then converted into a root wh- structure, after which high specificity elements are dropped to construct a more generic question (cf. Figure~\ref{fig:reformulation_example}, Q7,Q8, Q9). 
\OURMODEL-\ROOGEN only has an 8\% gap to the \ORACLE performance.
Figure~\ref{fig:answer_rate} shows that for all question types, \ROOGEN obtains the highest improvement in answer rate.

Comparing the answer rates of \ROOGEN and \ORACLE we make an interesting observation: although \ROOGEN combines all operators in $p$, its answer rate is still lower than \ORACLE.
This shows that applying all operators is not desirable for all questions. However, in practical settings, processing questions separately with all operators is not feasible due to the induced generation and QA latency. Hence, our proposed solution represents a trade-off between deployment feasibility and improvement in answer rate.

\subsection{Answer Relevance Results}\label{subsec:relevance}

It is important to consider if the provided answers to previously unanswered questions are relevant to the user's information need.
Since \OURMODEL performs numerous syntactic and semantic changes, there is a risk that the reformulated questions will result in answers that are not related to the user's intent. 

\Cref{t:human} shows the answer relevance results for the different operators based on a human study where answers are assessed for their relevance to $q$.

\REP has the highest exact relevance with 59\% (cf. Table~\ref{t:human}), but in absolute terms as shown in Table~\ref{rs:main} it obtains the lowest answer rate increase of 9.41\%.
The other operators are more complex and more likely to change the intent, the answer relevance is shifted towards \emph{related} and \emph{irrelevant} answers. For instance, \ROO and \GEN have the highest irrelevant answers, with 36\% and 29\%, respectively. This is intuitive given that the scope of the original question is reduced in $q'$, which can lead to unrelated answers.
On the other hand, we observe that \ROOGEN has the most answers that are related to $q$, with 73\% based on the human annotations, or 13.1\% on the 1M test set (extrapolated results).
It also obtains the least irrelevant answers as well as the highest answer rate, which we speculate is because the root wh- transformation and generalization reduce answering complexity and broaden recall, leading to a better pool of candidate answers for the QA system.
Furthermore, the different operators are complementary (cf. \Cref{sec:crosstabs}), hence, their combination achieves the best result.

\vspace{-1.5pt}
\subsection{Live QA Deployment}
\label{sec:deployment}

The \OURMODEL-\ROO model\footnote{We chose the best single operation model due to the doubled latency of \ROOGEN.} was deployed for real-time reformulation of unanswered questions in a leading voice assistant.
This live deployment enables answering for millions of previously unanswered requests.
Each day we solicit explicit binary relevance feedback from a portion of customers receiving answers of \OURMODEL reformulations,
with metrics exceeding or matching those reported in Table~\ref{t:human}.

\vspace{-1.5pt}
\section{Conclusion}

We tackled the problem of improving spoken QA, and analyzed questions from live data to identify key challenges that could be addressed with reformulation.
Based on this we proposed \OURMODEL with novel linguistically-motivated reformulation operators to solve the identified challenges.
Offline experiments show the effectiveness of our novel root transformation and generalization operations, with up to 24\% of unanswered questions being answered via reformulations with high answer relevance. Live deployment in a leading voice assistant has positively impacted millions of requests.

We showed reformulation helps QA systems adapt to spoken user questions. We presented key insights from a deployed solution showing that performance can be significantly increased, without changing the underlying QA backends, by simply improving questions in their syntax and semantics.

\section*{Limitations and Future Work}

In this work we did not consider the following aspects, which we discuss below and lay out directions for how to address them in future work.

\paragraph{Combining Reformulation Operations:} The reformulation operators, except \REP, which is applied jointly with other operators, are applied sequentially, in their given order, e.g. \ROOGEN. This has two potential limitations that we aim to address in future work. First, applying multiple operators sequentially has the negative impact of increased inference latency as the \OURMODEL model needs to be applied multiple times, which can become a bottleneck for systems that process large traffic volumes. Second,  by applying sequentially the reformulation operators, the likelihood of cascading errors or the model making mistakes in terms of the target reformulation shape increases.
We aim to address this limitation in the future by fine-tuning the model to jointly perform multiple reformulation operators in a single pass.

\paragraph{Large Language Models (LLM):} In this work we relied on BART~\cite{lewis-etal-2020-bart} as our seq2seq model, and did not experiment with newer multi-billion parameter LLMs. Recently we have seen rapid progress in the space of LLMs, both in terms of model size and their capabilities to perform various tasks~\cite{chung2022scaling}. However, we note that deploying LLMs is limited by their high inference latency, particularly in high-traffic, low-latency systems such as ours. Furthermore, for experimenting with API-based approaches such as ChatGPT and GPT-4, using these systems was not possible due to data confidentiality. While we will explore leveraging LLMs for this task in the future, current experimental results show that even smaller language models such as BART, with a sufficient amount of training data, can be fine-tuned to perform the task accurately.

\paragraph{Evaluation on Public Datasets:} Our evaluation focused on real-world unanswered user utterances from voice assistants. We did not use public datasets as currently available resources do not accurately represent customer behavior at scale. However, the community is aware of this divergence, and there are initial efforts in different NLP tasks to create public datasets that represent real-world user behavior. For example, in the the task of Named Entity Recognition there has been recent work on bridging the gap between academic datasets and real-world problems by creating new resources that represent contemporary challenges that are encountered in practice \cite{DBLP:journals/corr/abs-2305-06586,DBLP:conf/coling/MalmasiFFKR22}.
In future work we will consider evaluating \OURMODEL on such datasets as they become available. Furthermore, the findings from our work may be used to create data that includes the challenges we identified as part of our analysis (either by organically collecting such data, or simulating it to generate synthetic data).

\paragraph{Multilingual Experiments:} We only considered English-language questions in this work, and it will be of interest to consider how our approach can be extended to other languages using multilingual models. The evaluation of cross-lingual transfer for this task is another open research area.
\section*{Acknowledgments}
We would like to thank the following people for their valuable assistance and feedback: Danielle Class, Eugene Agichtein, Jim Fakonas, Leela Prabhu, Mohamed Nasreldin, and Zhiji Liu.

\bibliography{custom,internal}

\begin{thebibliography}{38}
\expandafter\ifx\csname natexlab\endcsname\relax\def\natexlab#1{#1}\fi

\bibitem[{Boldi et~al.(2011)Boldi, Bonchi, Castillo, and
  Vigna}]{boldi2011query}
Paolo Boldi, Francesco Bonchi, Carlos Castillo, and Sebastiano Vigna. 2011.
\newblock Query reformulation mining: models, patterns, and applications.
\newblock \emph{Information retrieval}, 14:257--289.

\bibitem[{Buck et~al.(2018)Buck, Bulian, Ciaramita, Gajewski, Gesmundo,
  Houlsby, and Wang.}]{buck2018ask}
Christian Buck, Jannis Bulian, Massimiliano Ciaramita, Wojciech Gajewski,
  Andrea Gesmundo, Neil Houlsby, and Wei Wang. 2018.
\newblock \href {https://openreview.net/forum?id=S1CChZ-CZ} {Ask the right
  questions: Active question reformulation with reinforcement learning}.
\newblock In \emph{International Conference on Learning Representations}.

\bibitem[{Byrd and Srivastava(2022)}]{byrd-srivastava-2022-predicting}
Matthew Byrd and Shashank Srivastava. 2022.
\newblock \href {https://doi.org/10.18653/v1/2022.acl-short.15} {Predicting
  difficulty and discrimination of natural language questions}.
\newblock In \emph{Proceedings of the 60th Annual Meeting of the Association
  for Computational Linguistics (Volume 2: Short Papers)}, pages 119--130,
  Dublin, Ireland. Association for Computational Linguistics.

\bibitem[{Chafe and Tannen(1987)}]{chafe1987relation}
Wallace Chafe and Deborah Tannen. 1987.
\newblock The relation between written and spoken language.
\newblock \emph{Annual review of anthropology}, 16(1):383--407.

\bibitem[{Chali and Hasan(2012)}]{chali-hasan-2012-simple}
Yllias Chali and Sadid~A. Hasan. 2012.
\newblock \href {https://aclanthology.org/W12-6001} {Simple or complex?
  classifying questions by answering complexity}.
\newblock In \emph{Proceedings of the Workshop on Question Answering for
  Complex Domains}, pages 1--10, Mumbai, India. The COLING 2012 Organizing
  Committee.

\bibitem[{Chen et~al.(2022)Chen, Zhao, Fang, Fetahu, Rokhlenko, and
  Malmasi}]{chen-etal-2022-reinforced}
Zhiyu Chen, Jie Zhao, Anjie Fang, Besnik Fetahu, Oleg Rokhlenko, and Shervin
  Malmasi. 2022.
\newblock \href {https://aclanthology.org/2022.emnlp-industry.36} {Reinforced
  question rewriting for conversational question answering}.
\newblock In \emph{Proceedings of the 2022 Conference on Empirical Methods in
  Natural Language Processing: Industry Track}, pages 357--370, Abu Dhabi, UAE.
  Association for Computational Linguistics.

\bibitem[{Choi et~al.(2021)Choi, Palomaki, Lamm, Kwiatkowski, Das, and
  Collins}]{choi-etal-2021-decontextualization}
Eunsol Choi, Jennimaria Palomaki, Matthew Lamm, Tom Kwiatkowski, Dipanjan Das,
  and Michael Collins. 2021.
\newblock \href {https://doi.org/10.1162/tacl_a_00377} {Decontextualization:
  Making sentences stand-alone}.
\newblock \emph{Transactions of the Association for Computational Linguistics},
  9:447--461.

\bibitem[{Chu et~al.(2020)Chu, Chen, Chen, Wang, Gimpel, Faruqui, and
  Si}]{chu-mqr-20}
Zewei Chu, Mingda Chen, Jing Chen, Miaosen Wang, Kevin Gimpel, Manaal Faruqui,
  and Xiance Si. 2020.
\newblock How to ask better questions? a large-scale multi-domain dataset for
  rewriting ill-formed questions.
\newblock In \emph{Proc. of {AAAI}}.

\bibitem[{Chung et~al.(2022)Chung, Hou, Longpre, Zoph, Tay, Fedus, Li, Wang,
  Dehghani, Brahma et~al.}]{chung2022scaling}
Hyung~Won Chung, Le~Hou, Shayne Longpre, Barret Zoph, Yi~Tay, William Fedus,
  Eric Li, Xuezhi Wang, Mostafa Dehghani, Siddhartha Brahma, et~al. 2022.
\newblock Scaling instruction-finetuned language models.
\newblock \emph{arXiv preprint arXiv:2210.11416}.

\bibitem[{Clark et~al.(2019)Clark, Lee, Chang, Kwiatkowski, Collins, and
  Toutanova}]{DBLP:conf/naacl/ClarkLCK0T19}
Christopher Clark, Kenton Lee, Ming{-}Wei Chang, Tom Kwiatkowski, Michael
  Collins, and Kristina Toutanova. 2019.
\newblock \href {https://doi.org/10.18653/v1/n19-1300} {Boolq: Exploring the
  surprising difficulty of natural yes/no questions}.
\newblock In \emph{Proceedings of the 2019 Conference of the North American
  Chapter of the Association for Computational Linguistics: Human Language
  Technologies, {NAACL-HLT} 2019, Minneapolis, MN, USA, June 2-7, 2019, Volume
  1 (Long and Short Papers)}, pages 2924--2936. Association for Computational
  Linguistics.

\bibitem[{Eck et~al.(2005)Eck, Vogel, and Waibel}]{eck-etal-2005-low}
Matthias Eck, Stephan Vogel, and Alex Waibel. 2005.
\newblock \href {https://aclanthology.org/2005.iwslt-1.7} {Low cost portability
  for statistical machine translation based on n-gram frequency and
  {TF}-{IDF}}.
\newblock In \emph{Proceedings of the Second International Workshop on Spoken
  Language Translation}, Pittsburgh, Pennsylvania, USA.

\bibitem[{Ennaciri(2022)}]{ennaciri2022searching}
Majda Ennaciri. 2022.
\newblock Searching for explanation of difficult scientific terms.

\bibitem[{Fader et~al.(2014)Fader, Zettlemoyer, and Etzioni}]{Fader14}
Anthony Fader, Luke Zettlemoyer, and Oren Etzioni. 2014.
\newblock {Open Question Answering Over Curated and Extracted Knowledge Bases}.
\newblock In \emph{KDD}.

\bibitem[{Fetahu et~al.(2023)Fetahu, Kar, Chen, Rokhlenko, and
  Malmasi}]{DBLP:journals/corr/abs-2305-06586}
Besnik Fetahu, Sudipta Kar, Zhiyu Chen, Oleg Rokhlenko, and Shervin Malmasi.
  2023.
\newblock \href {https://doi.org/10.48550/arXiv.2305.06586} {Semeval-2023 task
  2: Fine-grained multilingual named entity recognition (multiconer 2)}.
\newblock \emph{CoRR}, abs/2305.06586.

\bibitem[{Gan and Ng(2019)}]{gan-ng-2019-improving}
Wee~Chung Gan and Hwee~Tou Ng. 2019.
\newblock \href {https://doi.org/10.18653/v1/P19-1610} {Improving the
  robustness of question answering systems to question paraphrasing}.
\newblock In \emph{Proceedings of the 57th Annual Meeting of the Association
  for Computational Linguistics}, pages 6065--6075, Florence, Italy.
  Association for Computational Linguistics.

\bibitem[{Gupta et~al.(2021)Gupta, Xu, Upadhyay, Yang, and
  Faruqui}]{gupta2021disfl}
Aditya Gupta, Jiacheng Xu, Shyam Upadhyay, Diyi Yang, and Manaal Faruqui. 2021.
\newblock {Disfl-QA: A Benchmark Dataset for Understanding Disfluencies in
  Question Answering}.
\newblock In \emph{Findings of the Association for Computational Linguistics:
  ACL-IJCNLP 2021}, pages 3309--3319.

\bibitem[{He et~al.(2016)He, Tang, Ouyang, Kang, Yin, and
  Chang}]{he2016learning}
Yunlong He, Jiliang Tang, Hua Ouyang, Changsung Kang, Dawei Yin, and Yi~Chang.
  2016.
\newblock Learning to rewrite queries.
\newblock In \emph{Proceedings of the 25th ACM International on Conference on
  Information and Knowledge Management}, pages 1443--1452.

\bibitem[{Keskar et~al.(2019)Keskar, McCann, Varshney, Xiong, and
  Socher}]{keskar2019ctrl}
Nitish~Shirish Keskar, Bryan McCann, Lav~R Varshney, Caiming Xiong, and Richard
  Socher. 2019.
\newblock Ctrl: A conditional transformer language model for controllable
  generation.
\newblock \emph{arXiv preprint arXiv:1909.05858}.

\bibitem[{Kim et~al.(2021)Kim, Pavlick, Karagol~Ayan, and
  Ramachandran}]{kim-etal-2021-linguist}
Najoung Kim, Ellie Pavlick, Burcu Karagol~Ayan, and Deepak Ramachandran. 2021.
\newblock \href {https://doi.org/10.18653/v1/2021.acl-long.304} {Which linguist
  invented the lightbulb? presupposition verification for question-answering}.
\newblock In \emph{Proceedings of the 59th Annual Meeting of the Association
  for Computational Linguistics and the 11th International Joint Conference on
  Natural Language Processing (Volume 1: Long Papers)}, pages 3932--3945,
  Online. Association for Computational Linguistics.

\bibitem[{Ko et~al.(2020)Ko, Chen, Huang, Durrett, and
  Li}]{ko-etal-2020-inquisitive}
Wei-Jen Ko, Te-yuan Chen, Yiyan Huang, Greg Durrett, and Junyi~Jessy Li. 2020.
\newblock \href {https://doi.org/10.18653/v1/2020.emnlp-main.530} {Inquisitive
  question generation for high level text comprehension}.
\newblock In \emph{Proceedings of the 2020 Conference on Empirical Methods in
  Natural Language Processing (EMNLP)}, pages 6544--6555, Online. Association
  for Computational Linguistics.

\bibitem[{Lewis et~al.(2020)Lewis, Liu, Goyal, Ghazvininejad, Mohamed, Levy,
  Stoyanov, and Zettlemoyer}]{lewis-etal-2020-bart}
Mike Lewis, Yinhan Liu, Naman Goyal, Marjan Ghazvininejad, Abdelrahman Mohamed,
  Omer Levy, Veselin Stoyanov, and Luke Zettlemoyer. 2020.
\newblock \href {https://doi.org/10.18653/v1/2020.acl-main.703} {{BART}:
  Denoising sequence-to-sequence pre-training for natural language generation,
  translation, and comprehension}.
\newblock In \emph{Proceedings of the 58th Annual Meeting of the Association
  for Computational Linguistics}, pages 7871--7880, Online. Association for
  Computational Linguistics.

\bibitem[{Liu et~al.(2019)Liu, Zhang, Yan, Chang, and
  Yu}]{10.1145/3357384.3358046}
Ye~Liu, Chenwei Zhang, Xiaohui Yan, Yi~Chang, and Philip~S. Yu. 2019.
\newblock \href {https://doi.org/10.1145/3357384.3358046} {Generative question
  refinement with deep reinforcement learning in retrieval-based qa system}.
\newblock In \emph{Proceedings of the 28th ACM International Conference on
  Information and Knowledge Management}, CIKM '19, page 1643–1652, New York,
  NY, USA. Association for Computing Machinery.

\bibitem[{Malmasi et~al.(2022)Malmasi, Fang, Fetahu, Kar, and
  Rokhlenko}]{DBLP:conf/coling/MalmasiFFKR22}
Shervin Malmasi, Anjie Fang, Besnik Fetahu, Sudipta Kar, and Oleg Rokhlenko.
  2022.
\newblock \href {https://aclanthology.org/2022.coling-1.334} {Multiconer: {A}
  large-scale multilingual dataset for complex named entity recognition}.
\newblock In \emph{Proceedings of the 29th International Conference on
  Computational Linguistics, {COLING} 2022, Gyeongju, Republic of Korea,
  October 12-17, 2022}, pages 3798--3809. International Committee on
  Computational Linguistics.

\bibitem[{Martin et~al.(2020)Martin, de~la Clergerie, Sagot, and
  Bordes}]{martin-etal-2020-controllable}
Louis Martin, {\'E}ric de~la Clergerie, Beno{\^\i}t Sagot, and Antoine Bordes.
  2020.
\newblock \href {https://aclanthology.org/2020.lrec-1.577} {Controllable
  sentence simplification}.
\newblock In \emph{Proceedings of the Twelfth Language Resources and Evaluation
  Conference}, pages 4689--4698, Marseille, France. European Language Resources
  Association.

\bibitem[{Mielke et~al.(2019)Mielke, Cotterell, Gorman, Roark, and
  Eisner}]{mielke-etal-2019-kind}
Sabrina~J. Mielke, Ryan Cotterell, Kyle Gorman, Brian Roark, and Jason Eisner.
  2019.
\newblock \href {https://doi.org/10.18653/v1/P19-1491} {What kind of language
  is hard to language-model?}
\newblock In \emph{Proceedings of the 57th Annual Meeting of the Association
  for Computational Linguistics}, pages 4975--4989, Florence, Italy.
  Association for Computational Linguistics.

\bibitem[{Moradlou et~al.(2021)Moradlou, Zheng, Ye, and
  Ginzburg}]{moradlou2021wh}
Sara Moradlou, Xiaobei Zheng, TIAN Ye, and Jonathan Ginzburg. 2021.
\newblock Wh-questions are understood before polar-questions: Evidence from
  english, german, and chinese.
\newblock \emph{Journal of Child Language}, 48(1):157--183.

\bibitem[{Motro(1984)}]{motro1984query}
Amihai Motro. 1984.
\newblock Query generalization: A method for interpreting null answers.
\newblock In \emph{Expert Database Workshop}, pages 597--616.

\bibitem[{Nassar et~al.(2019)Nassar, Ananda-Rajah, and
  Haffari}]{nassar-etal-2019-neural}
Islam Nassar, Michelle Ananda-Rajah, and Gholamreza Haffari. 2019.
\newblock \href {https://aclanthology.org/U19-1023} {Neural versus non-neural
  text simplification: A case study}.
\newblock In \emph{Proceedings of the The 17th Annual Workshop of the
  Australasian Language Technology Association}, pages 172--177, Sydney,
  Australia. Australasian Language Technology Association.

\bibitem[{Pomerantz(2005)}]{pomerantz2005linguistic}
Jeffrey Pomerantz. 2005.
\newblock A linguistic analysis of question taxonomies.
\newblock \emph{Journal of the American Society for Information Science and
  Technology}, 56(7):715--728.

\bibitem[{Raffel et~al.(2020)Raffel, Shazeer, Roberts, Lee, Narang, Matena,
  Zhou, Li, and Liu}]{2020t5}
Colin Raffel, Noam Shazeer, Adam Roberts, Katherine Lee, Sharan Narang, Michael
  Matena, Yanqi Zhou, Wei Li, and Peter~J. Liu. 2020.
\newblock \href {http://jmlr.org/papers/v21/20-074.html} {Exploring the limits
  of transfer learning with a unified text-to-text transformer}.
\newblock \emph{Journal of Machine Learning Research}, 21(140):1--67.

\bibitem[{Riezler and Liu(2010)}]{riezler-liu-2010-query}
Stefan Riezler and Yi~Liu. 2010.
\newblock \href {https://doi.org/10.1162/coli_a_00010} {Query rewriting using
  monolingual statistical machine translation}.
\newblock \emph{Computational Linguistics}, 36(3):569--582.

\bibitem[{Salesky et~al.(2019)Salesky, Sperber, and Waibel}]{salesky2019fluent}
Elizabeth Salesky, Matthias Sperber, and Alex Waibel. 2019.
\newblock {Fluent Translations from Disfluent Speech in End-to-End Speech
  Translation}.
\newblock In \emph{Proceedings of NAACL}, pages 2786--2792.

\bibitem[{Sheang and Saggion(2021)}]{sheang-saggion-2021-controllable}
Kim~Cheng Sheang and Horacio Saggion. 2021.
\newblock \href {https://aclanthology.org/2021.inlg-1.38} {Controllable
  sentence simplification with a unified text-to-text transfer transformer}.
\newblock In \emph{Proceedings of the 14th International Conference on Natural
  Language Generation}, pages 341--352, Aberdeen, Scotland, UK. Association for
  Computational Linguistics.

\bibitem[{Shriberg(2005)}]{shriberg2005spontaneous}
Elizabeth Shriberg. 2005.
\newblock Spontaneous speech: How people really talk and why engineers should
  care.
\newblock In \emph{Ninth European Conference on Speech Communication and
  Technology}.

\bibitem[{Tomuro(2003)}]{tomuro-2003-interrogative}
Noriko Tomuro. 2003.
\newblock \href {https://doi.org/10.3115/1118984.1118989} {Interrogative
  reformulation patterns and acquisition of question paraphrases}.
\newblock In \emph{Proceedings of the Second International Workshop on
  Paraphrasing}, pages 33--40, Sapporo, Japan. Association for Computational
  Linguistics.

\bibitem[{Ward(1989)}]{ward-1989-understanding}
Wayne Ward. 1989.
\newblock \href {https://aclanthology.org/H89-1018} {Understanding spontaneous
  speech}.
\newblock In \emph{Speech and Natural Language: Proceedings of a Workshop Held
  at Philadelphia, {P}ennsylvania, {F}ebruary 21-23, 1989}.

\bibitem[{Yang et~al.(2018)Yang, Qi, Zhang, Bengio, Cohen, Salakhutdinov, and
  Manning}]{yang-etal-2018-hotpotqa}
Zhilin Yang, Peng Qi, Saizheng Zhang, Yoshua Bengio, William Cohen, Ruslan
  Salakhutdinov, and Christopher~D. Manning. 2018.
\newblock \href {https://doi.org/10.18653/v1/D18-1259} {{H}otpot{QA}: A dataset
  for diverse, explainable multi-hop question answering}.
\newblock In \emph{Proceedings of the 2018 Conference on Empirical Methods in
  Natural Language Processing}, pages 2369--2380, Brussels, Belgium.
  Association for Computational Linguistics.

\bibitem[{Ye et~al.(2022)Ye, Ng, and Han}]{ye2022robustness}
Hai Ye, Hwee~Tou Ng, and Wenjuan Han. 2022.
\newblock {On the Robustness of Question Rewriting Systems to Questions of
  Varying Hardness}.
\newblock In \emph{Proceedings of the 60th Annual Meeting of the Association
  for Computational Linguistics (Volume 1: Long Papers)}, pages 2100--2113.

\end{thebibliography}
\bibliographystyle{acl_natbib}

\appendix
\crefalias{section}{appendix}
\section*{\centering Appendix}

\section{Spoken QA Failure Analysis}
\label{sec:failure_analysis}

We analyzed data to understand prevalent challenges in spoken QA failure. 
Unlike prior work, which uses Machine Reading Comprehension (MRC) datasets like SQuAD, we leverage real questions from a leading voice assistant.\footnote{We do not consider ASR challenges in this work, and only deal with text transcripts.}

We performed a quantitative analysis, taking two large random samples of answered and unanswered user queries, totalling 10 million unique questions.
For all questions, we compute several sentence-level variables (length, type-token ratio, TF-IDF) which are predictive of language complexity \cite{mielke-etal-2019-kind,byrd-srivastava-2022-predicting,ennaciri2022searching},
and measure their correlation with whether the question was answered.
We also hypothesize that the question's linguistic shape is important (see \S\ref{sec:related_work}). Following prior work \cite{pomerantz2005linguistic}, we define a syntactic question typology (\Cref{tab:question_types}) and develop an accurate type classification heuristic (c.f. \Cref{sec:qtype_heuristic} for details).

\begin{table}[!hbt]
\centering
    \resizebox{1.0\columnwidth}{!}{
        \begin{tabular}{p{2.82cm} p{7cm}}\toprule
        \textbf{Type} & \textbf{Description} \\
        \hline
        \textbf{Root wh-question}        & The wh-phrase is clause-initial. (\exampleb{Who is the US president?}. \exampleb{How large is an elephant?}) \\ \hline
        \textbf{Polar (Yes-No)}  & Asks if a statement is true. (e.g. \exampleb{Is it going to rain tomorrow?}, \exampleb{Can cats eat onions?}) \\ \hline
        \textbf{Open}            & Open-ended \textit{how} questions.  (e.g. \exampleb{How does depression affect the body?}) \\
        \hline
        \textbf{Request}         & Direct request beginning with a verb. (e.g. \exampleb{Tell me the capital of Utah.}) \\
        \hline
        \textbf{Other}           & Any other utterance. (e.g. \exampleb{watermelon health benefits}, \exampleb{sports softball in Denver}) \\
        \bottomrule
        \end{tabular}}
    \caption{\textmd{A list of the types in our question typology.}}
    \label{tab:question_types}
\end{table}

\begin{table}[ht!]
\centering
\resizebox{\columnwidth}{!}{
\begin{tabular}{lr}
\toprule
\textbf{Variable} & \multicolumn{1}{r}{\textbf{Pearson Correlation ($r$)}} \\ \midrule
Token Length & $-0.25$\\
Char Length & $-0.24$ \\
Type-Token Ratio (TTR) & $+0.12$ \\
Mean of IDF scores & $-0.13$ \\
Sum of IDF scores & $-0.30$ \\
Sum of TF-IDF scores & $-0.31$ \\
Mean of TF-IDF scores & $-0.12$ \\
\midrule
\textbf{Question Type} & \textbf{Difference w/ Answered} \\ \midrule
Root & $-10.4$\% \\
Polar & $+8.5$\% \\
Open & $+2.4$\% \\
Request & $-2.1$\% \\
Other & $+1.6$\% \\
\bottomrule
\end{tabular}}
\caption{\textmd{Top: Pearson correlation between question characteristics and if it was answered (all $p < 0.001$). Bottom: Distributional differences in question types between the unanswered and answered questions.}} 
\label{tab:question_analysis}
\end{table}

\Cref{tab:question_analysis} shows the results.
For confidentiality, we only report correlations and relative differences between the answered and unanswered groups, whose sizes cannot be disclosed.
We note that longer questions and those with higher specificity (\ie IDF) are more likely to be unanswered. Higher TTR (\ie fewer repeated tokens) results in higher answer rates, likely because repetition is associated with disfluencies.
Question type also has a big impact on answerability. Simple root wh-questions are less prevalent in the answered subset, while polar questions are much more frequent in the unanswered subset.

\section{Question Type Classification}
\label{sec:qtype_heuristic}

We develop a rule-based algorithm to classify a question into a predefined type (cf. Table \ref{tab:question_types}).

Algorithm~\ref{alg:heuristic} shows our heuristic to determine the question type. The algorithm is a rule-based and applied in cascade,
until there is a match between question and type. The evaluation order is the same as listed in Table \ref{tab:question_types}, from top to bottom.

\begin{itemize}[leftmargin=*]
    \itemsep0em
    \item \textbf{Root}: a question which starts with a \emph{wh-*} or some specific \emph{how-} bigrams.
    \item \textbf{Polar}: A yes or no question starting with predefined keywords.
    \item \textbf{Open}: Start with \emph{how}, but not a root question.
    \item \textbf{Request}: They are a command to a QA system and start with a verb.\footnote{We use spaCy for POS-Tagging.}
    \item \textbf{Other}: If anything else, sentences are labeled as other.
\end{itemize}

Below are listed some of the input variables necessary for Algorithm~\ref{alg:heuristic}.
\begin{itemize}[leftmargin=*]
    \itemsep0em
    \item \textbf{wh-* or how-* bigrams}: \example{what}, \example{where}, \example{when}, \example{which}, \example{who}, \example{why}, \example{how much}, \example{how many}, \example{how long}, \example{how old}, \example{how early}, \example{how soon}, \example{how wealthy}, \example{how rich}, \example{how big}, \example{how small}, \example{how tall}, \example{how short}, \example{how heavy}, \example{how often}, \example{how late}, \example{how far}, \example{how high}, \example{how fast}, \example{how quickly}, \example{how close};
    \item \textbf{polar keywords}: \example{do}, \example{does}, \example{did}, \example{can}, \example{was}, \example{were}, \example{should}, \example{is }, \example{isn}, \example{has}, \example{have}, \example{are}, \example{aren}, \example{will};
\end{itemize}

\begin{algorithm}
\caption{Heuristic for question types}\label{alg:heuristic}
\begin{algorithmic}
\Require sentence $ s $

\If{$s$ starts with wh-* or how-* bigrams} 
    \State type $\gets$ root
\ElsIf{$s$ starts with a keyword from polar keywords list}
    \State type $ \gets $ polar
\ElsIf{$s$ starts with ``how''}
    \State type $ \gets $ open 
\ElsIf{$s$ starts with verb}
    \State type $ \gets $ request 
\Else
    \State type $\gets$ other
\EndIf
\State \Return type
\end{algorithmic}
\end{algorithm}

\subsection{Heuristic Accuracy Evaluation}
To evaluate the accuracy of our heuristic algorithm, we randomly sampled 100 questions from each question type from the testing set and annotated whether the classified question type is correct. In total, 500 questions were annotated and the overall accuracy is 95\%. The accuracy of each question type is summarized in Table~\ref{t:q-acc}.

\begin{table}[!bht]
\centering
\begin{tabular}{lc}
\toprule
\textbf{Question Type} & \textbf {Accuracy (\%)} \\ \midrule
Root                   & $0.98$              \\
Polar                  & $0.98$              \\
Open                   & $0.95$              \\
Request                & $0.89$              \\
Other                  & $0.95$              \\ 
\midrule
Average      &    $0.95$ \\
\bottomrule
\end{tabular}
\caption{Question classification heuristic accuracy (based on human assessment), for each question type.}\label{t:q-acc}
\end{table}

\begin{table}[]
\begin{tabular}{lrr}
\toprule
\textbf{Original Question Type} & \textbf{\ROO} & \textbf{\REP} \\ \midrule
Root & 0 & 544,440 \\
Polar & 23,201 & 200,942 \\
Open & 36,322 & 257,704 \\
Request & 15,405 & 450 \\
Other & 113,568 & 9,348 \\ \midrule
Total & 188,496 & 1,012,884 \\ \bottomrule
\end{tabular}
\caption{Distributions of question types in the weakly supervised pre-training data for the \ROO and \REP operators.}
\label{t:pretrain_dist}
\end{table}

\begin{table*}[]
\resizebox{1.0\textwidth}{!}{
\centering
\begin{tabular}{llll}
\toprule
\multicolumn{1}{c}{\textbf{Question}} & \textbf{Source Type} & \multicolumn{1}{c}{\textbf{Reformulation}} & \textbf{Target Type} \\ \midrule
\example{where does spider live in?} & root & \example{where does a spider live?} & root \\
\example{what is the oridgin of the word mosque?} & root & \example{where does the word mosque come from?} & root \\
\example{how remember pronunciation of danish words?} & open & \begin{tabular}[c]{@{}l@{}}\emph{``how can i remember the pronunciation} \\ \emph{of danish words?''}\end{tabular} & root \\
\example{how can we make money from youtube?} & open & \example{how do people earn money from youtube?} & root \\
\example{does the grammar generates the words?} & polar & \example{does the grammar generate the words?} & polar \\
\example{can charity claim patent on medicine?} & polar & \example{can charities be granted patents on medicine?} & polar \\
\example{winners in olympic in 2000?} & other & \example{names of olympic winners of 2008?} & other \\
\example{at what tempature does alcohol freeze?} & other & \example{at what temperture does alcohol freeze?} & other \\
\example{find out some advantages for setting up a partnership?} & request & \example{give 2 advantages of a business partnership?} & request \\
\begin{tabular}[c]{@{}l@{}}\emph{``name three groups of polymers and}\\ \emph{name one type of a composite?''}\end{tabular} & request & \example{name three common polymers?} & request \\ \bottomrule
\end{tabular}}
\caption{\REP examples of weakly-labeled pre-training data from the MQR dataset as labeled by our heuristics.}\label{t:pretrain-rep}
\end{table*}

\begin{table*}[]
\resizebox{1.0\textwidth}{!}{
\centering
\begin{tabular}{llll}
\toprule
\multicolumn{1}{c}{\textbf{Question}} & \multicolumn{1}{c}{\textbf{Source Type}} & \multicolumn{1}{c}{\textbf{Reformulation}} & \multicolumn{1}{c}{\textbf{Target Type}} \\ \midrule
\begin{tabular}[c]{@{}l@{}} \emph{``how do you know if your local bike} \\ \emph{club is worth paying for?''}\end{tabular} & open & \example{what benefits do bike clubs provide?} & root \\
\example{how do you forgive other people?} & open & \example{what's the best way to forgive people?} & root \\
\example{an example of enzyme mimic is ?} & other & \example{what are examples of enzymes and antibodies ?} & root \\
\begin{tabular}[c]{@{}l@{}}\emph{``basic difference between compilers}\\  \emph{and interpreters?''}\end{tabular} & other & \begin{tabular}[c]{@{}l@{}}\emph{``what are the differences between compilers}\\  \emph{and interpreters?''}\end{tabular} & root \\
\example{explain the ending of batman arkham city to me} & request & \example{what happens in the ending of batman arkham city?} & root \\
\example{unscrewing sliding window lock} & request & \begin{tabular}[c]{@{}l@{}}\emph{``what tool works with this star-shaped screw}\\  \emph{with a post in the middle?''}\end{tabular} & root \\
\example{do blackholes exist?} & polar & \example{why do black holes exist?} & root \\
\example{is there any easy way to make money online?} & polar & \example{what is the easiest way to earn money from online?} & root \\
\example{are there any good challenging puzzles?} & polar & \example{what are some good word puzzles?} & root \\ \bottomrule
\end{tabular}}
\caption{\ROO examples of weakly-labeled pre-training data from the MQR dataset as labeled by our heuristics.}\label{t:pretrain-roo}
\end{table*}

\pagebreak

\section{Model Implementation Details}
\label{subsec:implementation_details}
For both our approach and the baseline, we adopt BART~\cite{lewis-etal-2020-bart}\footnote{\url{https://huggingface.co/facebook/bart-base}} 
as our reformulation model $\mathcal{F}$. 
As annotating the \GEN task is not possible for all questions (as not all of them are generalizable, \eg \example{Who is Joe Biden?}), this results in a smaller amount of training data for the \GEN task. To address this, we upsampled the generalized reformulations by 5x during training so that the number of generalization samples matches other types of reformulations.
We train it for up to 20 epochs with a learning rate of $lr=1e-6$ and use Adam as our optimizer, and batch size of 16. The training is halted using early stopping, if the validation loss is non-decreasing after 3 epochs.

\section{Pre-training Data}
\label{sec:pretrain}

To prepare the weakly-supervised data for pre-training, we first apply our question type heuristic from \Cref{sec:qtype_heuristic} to classify the original questions and reformulations in the MQR dataset.\footnote{\url{https://github.com/ZeweiChu/MQR}}
We then automatically derive operator task labels from those question types using the method described in \S\ref{sec:data}.
This process yields $1.2$M samples. \Cref{t:pretrain_dist} shows the distribution of task labels and question types in the data. As noted earlier, data for the \GEN operator cannot be reliably derived with weak supervision on this dataset. The large majority of the data contains repairs, as that is the intended purpose of the MQR dataset.
Table~\ref{t:pretrain-rep} and Table~\ref{t:pretrain-roo} list some example questions from the MQR dataset with their assigned question types for the \REP and \ROO operators, respectively.

\section{Annotation Guidelines}
\label{sec:sop}
Here we describe in detail the question reformulation annotation guidelines. First, the steps for each reformulation operator are described, then a general overview of annotation guidelines for the entire annotation process is shown.

\subsection{Instructions for \REP}\label{sop:rep}
\REP reformulations must:%
\begin{itemize}[leftmargin=*]
    \itemsep0em
    \item not contain repetitions, false starts, and self corrections.
    \item be grammatically correct. For example, \example{Is Bill Pullman have a son?} $\rightarrow$ \example{Does Bill Pullman have a son?}. %
    \item be impersonal and formal. For example, \example{Where can I get a booze after 11 pm?} $\rightarrow$ \example{Which stores sell beer after 11 pm?}.
    \item keep the original question type (\eg root $\rightarrow$ root). %
\end{itemize}

\subsection{Instructions for \ROO}\label{sop:roo}

\ROO question reformulations must satisfy the following constraints:

\begin{itemize}[leftmargin=*]
    \itemsep0em 
    \item The reformulation must be a root question as defined in \Cref{sec:qtype_heuristic}. For example, \example{Is there any easy way to make money online?} $\rightarrow$ \example{What is the easiest way to earn money online?}.
    \item Reformulations must retain the intent of the original question. In the above example, the question type is changed from polar to root. However, the answer to the reformulated root question can still provide an answer to the original polar question. Reformulations where the intent is changed are invalid: \example{Can you freeze chicken that's already been thawed?} $\rightarrow$ \example{How long can chicken be frozen for before going bad?}.
    \item The reformulation additionally should satisfy the \REP constraints, with the exception of altering the question type.
\end{itemize}

\subsection{Instructions for \GEN}\label{sop:gen}
\GEN reformulations may slightly change the information that is sought in the original question to something more general.
This can be done by removing parts of a question (adjuncts or other clauses), and modifying referenced entities. 
Note that we do not make parallel entity changes (e.g. \example{Los Angeles} $\rightarrow$ \example{San Francisco}), but rather perform vertical generalization (\eg with hypernyms or holonyms \example{Los Angeles} $\rightarrow$ \example{California}). There are different cases to generalize a question:
\begin{itemize}[leftmargin=*]
    \itemsep0em
    \item The reformulation is less restricted than the original question w.r.t some entity (\eg \example{What do pythons eat?} $\rightarrow$ \example{What do snakes eat?});
    \item The reformulation is more general than the original question regarding conditions/constraints (\eg \example{Who is the tallest person in the USA?} $\rightarrow$ \example{Who is the tallest person?}).
\end{itemize}

For any given question, multiple distinct generalizations may be possible.

\subsection{Overall Guidelines for Annotators}

You will be given questions and asked to generalize them or reshape them into other types.  All your reformulations must be done with respect to the original question. An original question can be generalized up to 3 times. 
Please complete the following steps for each question:
\begin{itemize}[leftmargin=*]
    \itemsep0em
    \item Question Validity (prior to any reformulation):
    \begin{enumerate}
        \item Judge whether the question seeks a valid answer. A question is invalid if you are unable to understand the question's intent. Or, alternatively, you judge that the question is unanswerable. This may be the case for personal questions (e.g. \example{do I have COVID?}). If the question is invalid, remove question from the training dataset.
    \end{enumerate}
    \item Perform \REP reformulation: 
    \begin{enumerate}
        \item Refer to \S\ref{sop:rep} to make sure your reformulation adheres to all \REP constraints.
    \end{enumerate}
    \item Do \ROO reformulation: 
    \begin{enumerate}
        \item Refer to \S\ref{sop:roo} to make sure your reformulation adheres to all constraints.
        \item If it is unfeasible to make the reformulation without changing the question’s intent, leave blank.
        \item Do not reformulate root questions.
    \end{enumerate}
    \item Do \GEN reformulation: 
    \begin{enumerate}
        \item Write down up to 3 generalized reformulation of the original question. If possible, try to perform different types of generalization.
        \item Refer to \S\ref{sop:gen} to make sure your reformulation adheres to all constraints.
    \end{enumerate}
\end{itemize}

\subsection{Sampling Strategy}\label{apdx:sample}

To sample questions for annotation, we first filter questions with fewer than 5 tokens or more than 13 tokens. Then we adopt the unseen strategy~\cite{eck-etal-2005-low} using bi-grams to select questions that cover diverse topics. For each question, we collect up to 3 different generalized reformulations, given that a question can be generalized in different ways. %

\section{Extrinsic Evaluation Data}\label{apdx:eval}
To evaluate the performance of reformulations on our QA system, we take a representative sample of 1M unanswered questions from the real traffic as the test set,  where the distribution across different question types is kept to the real traffic distribution. However, due to confidentially reasons, we cannot reveal the exact question type distribution.%

\section{Operator Contingency Tables}
\label{sec:crosstabs}

A natural question is whether different operators are correlated, \ie they lead to improved answering on the same set of questions, or if they are complementary/orthogonal by improving non-overlapping subsets of questions. To understand this relationship we performed a cross tabulation analysis by building $2$x$2$ contingency tables comparing different operators on our test set. Each operator is represented by a binary variable indicating whether the reformulation by that operator resulted in the unanswered question becoming answered.

\Cref{t:crosstab} shows the results of this analysis for the \OURMODEL model.
We observe that there is substantial degree of orthogonality between the operators, as evidenced by cases where one operator fails and the other succeeds, \eg \ROO can improve answering on $6.98\%$ of the data where \REP fails to do so. The largest correlation is between \ROO and \ROOGEN, while the lowest is between \ROO and \REP.
All operators are best complemented by \ROOGEN.
The trend is inline with the results shown in Table~\ref{rs:main}, where \ROOGEN has the highest number of answered reformulated questions.

\begin{table*}[!hbt]
\centering
\begin{tabular}{crrccrr}
\cline{1-3} \cline{5-7}
\multicolumn{1}{|c|}{\textbf{}} & \multicolumn{1}{c}{\textbf{\REP= 0}} & \multicolumn{1}{c|}{\textbf{\REP= 1}} & \multicolumn{1}{c|}{} & \multicolumn{1}{c|}{} & \multicolumn{1}{c}{\textbf{\ROOGEN= 0}} & \multicolumn{1}{c|}{\textbf{\ROOGEN= 1}} \\ \cline{1-3} \cline{5-7} 
\multicolumn{1}{|c|}{\textbf{\GEN= 0}} & 83.80\% & \multicolumn{1}{r|}{1.87\%} & \multicolumn{1}{c|}{} & \multicolumn{1}{c|}{\textbf{\GEN= 0}} & 77.81\% & \multicolumn{1}{r|}{7.85\%} \\
\multicolumn{1}{|c|}{\textbf{\GEN= 1}} & 6.79\% & \multicolumn{1}{r|}{7.54\%} & \multicolumn{1}{c|}{} & \multicolumn{1}{c|}{\textbf{\GEN= 1}} & 4.27\% & \multicolumn{1}{r|}{10.07\%} \\ \cline{1-3} \cline{5-7} 
\multicolumn{1}{|c|}{\textbf{\ROO= 0}} & 83.61\% & \multicolumn{1}{r|}{3.21\%} & \multicolumn{1}{c|}{} & \multicolumn{1}{c|}{\textbf{\ROO= 0}} & 80.28\% & \multicolumn{1}{r|}{6.55\%} \\
\multicolumn{1}{|c|}{\textbf{\ROO= 1}} & 6.98\% & \multicolumn{1}{r|}{6.20\%} & \multicolumn{1}{c|}{} & \multicolumn{1}{c|}{\textbf{\ROO= 1}} & 1.80\% & \multicolumn{1}{r|}{11.38\%} \\ \cline{1-3} \cline{5-7} 
 & \multicolumn{1}{c}{} & \multicolumn{1}{c}{} &  &  & \multicolumn{1}{c}{} & \multicolumn{1}{c}{} \\ \cline{1-3} \cline{5-7} 
\multicolumn{1}{|c|}{} & \multicolumn{1}{c}{\textbf{\GEN= 0}} & \multicolumn{1}{c|}{\textbf{\GEN= 1}} & \multicolumn{1}{c|}{} & \multicolumn{1}{c|}{} & \multicolumn{1}{c}{\textbf{\ROOGEN= 0}} & \multicolumn{1}{c|}{\textbf{\ROOGEN= 1}} \\ \cline{1-3} \cline{5-7} 
\multicolumn{1}{|c|}{\textbf{\ROO= 0}} & 80.27\% & \multicolumn{1}{r|}{6.55\%} & \multicolumn{1}{c|}{} & \multicolumn{1}{c|}{\textbf{\REP= 0}} & 79.25\% & \multicolumn{1}{r|}{11.34\%} \\
\multicolumn{1}{|c|}{\textbf{\ROO= 1}} & 5.39\% & \multicolumn{1}{r|}{7.79\%} & \multicolumn{1}{c|}{} & \multicolumn{1}{c|}{\textbf{\REP= 1}} & 2.83\% & \multicolumn{1}{r|}{6.58\%} \\ \cline{1-3} \cline{5-7} 
\end{tabular}
\caption{Comparing the percentage of answered (1) and unanswered (0) questions between two operations in crosstables.}\label{t:crosstab}
\end{table*}

\section{Analysis of Reformulations Changes}
\label{sec:change_analysis}

We also consider how our reformulation operators change the original questions in terms of length and type-token ratio (TTR).
Previously, in Table~\ref{tab:question_analysis} of \Cref{sec:failure_analysis} we showed that these question characteristics are correlated with answer rate.
As a follow up, we examined how the \OURMODEL reformulations change these variables.

\begin{figure}[t!] %
\includegraphics[width=\columnwidth]{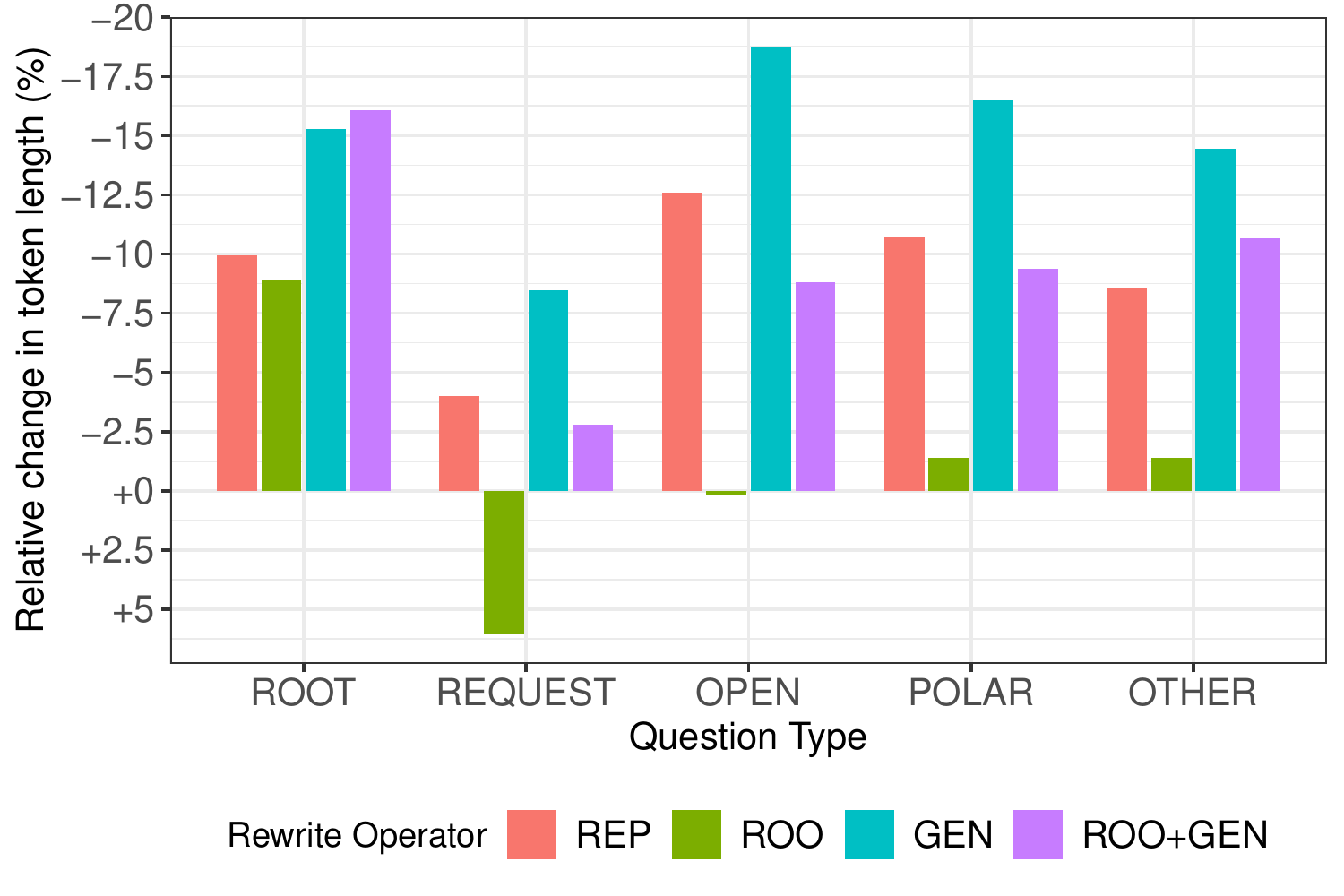}
\caption[]{Relative change in token length after applying the different reformulation operators.}
\label{fig:delta_tk}
\end{figure}

\begin{figure}[t!] %
\includegraphics[width=\columnwidth]{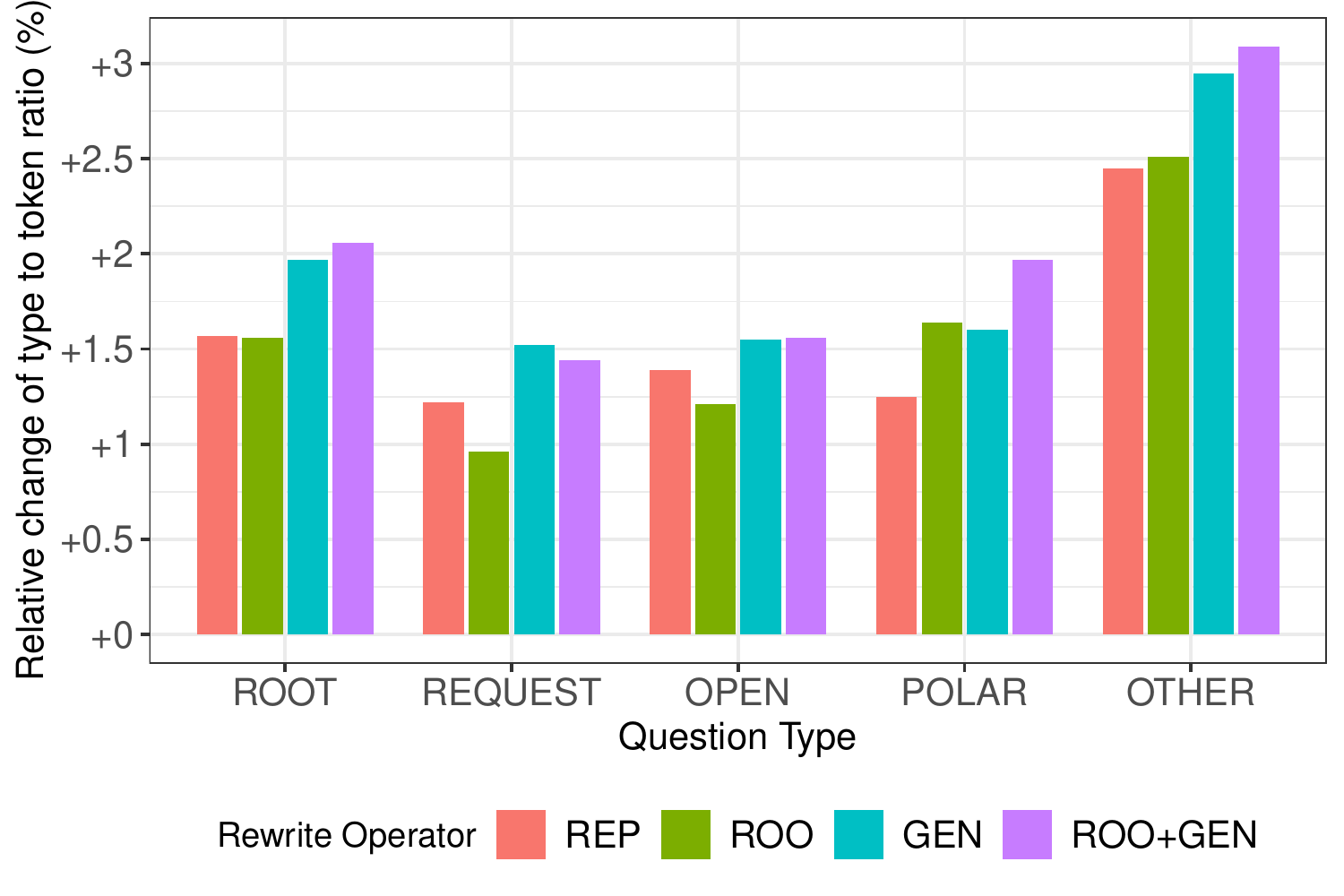}
\caption[]{Relative increase in type-token ratio after applying the different reformulation operators. }
\label{fig:delta_ttr}
\end{figure}

\Cref{fig:delta_tk} shows that \OURMODEL reformulations from all operators significantly shorten the input questions, indicating that they result in simplified questions.
The micro-averaged length reduction across all question types for each operator is $9.9\%$ for \REP, $4.7\%$ for \ROO, $15.5\%$ for \GEN, and $12.6\%$ for \ROOGEN.
The average length of a question reformulation by \ROO increases only for open and request question types, while it decreases in all other cases.
However, for open and request question types, \ROO makes the question more specific (\eg, \example{explain how to play football} is reformulated into \example{what is the best way to play football?} by \ROO). Sometimes, \ROO also makes polar question more specific (\eg, \example{is aspartame safe?} $\rightarrow$ \example{what are the health risks of eating aspartame?}).

Regarding TTR, from \Cref{fig:delta_ttr} we observe a consistent TTR increase for all reformulation types.
This is consistent with the findings from \Cref{tab:question_analysis}, where it was shown that TTR is positively correlated with answer rate. The increase of TTR means the reformulated questions are less likely to have repeated tokens.

\section{Question Reformulation Examples}
\label{sec:examples}

We show synthetic examples of question reformulations output by our model in \Cref{apdx:examples}.

\begin{table*}[]
\centering
\resizebox{1.0\textwidth}{!}{
\begin{tabular}{r p{11cm}p{8.1cm}}
\toprule
\textbf{Task} & \multicolumn{1}{c}{\textbf{Origin}} & \multicolumn{1}{c}{\textbf{Reformulation}} \\ 
\midrule
\REP & \example{why why are the dinosaurs go away from them} & \example{why did dinosaurs go away?} \\
\REP & \example{how do tiger how do baby tigers speak} & \example{how to communicate with baby tigers?}\\
\REP & \example{korean food new jersey} & \example{what is the best korean food in new jersey} \\
\REP & \example{hey find me some vinegar potato salad recipes} & \example{how to make vinegar potato salad} \\
\REP & \example{are bill pullman have a son} & \example{does bill pullman have a son} \\
\REP & \example{what time of the year do all the lightning bugs come out} & \example{when do lightning bugs come out} \\
\REP & \example{when is new movie about minions} & \example{when will the next movie about minions be released} \\
\REP & \example{how much does it how long does it take to boil one and a half cups of water} & \example{how long does it take to boil water}\\
\REP & \example{what's the macros of rotisserie chicken} & \example{what are the macros of a rotisserie chicken} \\
\midrule
\ROO & \example{composer dona zetti} & \example{who is the composer dona zetti} \\
\ROO & \example{does a lot of sugar will hurt your teeth} & \example{why does sugar hurt your teeth} \\
\ROO & \example{hey find me vinegar potato salad recipes} & \example{what are some good vinegar potato salad recipes?} \\
\ROO & \example{look up some macron recipes} & \example{what are some macron recipes} \\
\ROO & \example{hey show me a marinade for pepper steak} & \example{what marinade should be used for pepper steak} \\
\ROO & \example{are bill pullman have a son} & \example{who is bill pullman's son} \\
\ROO & \example{in an air fryer at what temperatures do i heat fully cooked bratwurst and for how long}  & \example{how long should bratwurst be cooked in an air fryer} \\ 
\ROO & \example{yes did steven spielberg go to films school} & \example{where did steven spielberg go to film school} \\
\ROO & \example{how do you do reminder for my phone} & \example{how to set up a reminder for a phone?} \\
\ROO & \example{again show me the recipe for pan seared steak} & \example{what are some recipes for pan seared steak} \\
\midrule
\GEN & \example{how long can fleas live without host} & \example{how long can fleas live} \\
\GEN & \example{in 1901 who won the first nobel prize for physics} & \example{who won the first nobel prize for physics} \\
\GEN & \example{are large feet assigned beauty some cultures}  & \example{are large feet assigned beauty} \\
\GEN & \example{did kamala harris ever move to canada than back to america}  & \example{did kamala harris move to canada} \\

\GEN & \example{how many playable characters are in lego star wars the skywalker saga} & \example{how many playable characters are in lego star wars} \\
\GEN & \example{tell me what is the total number of people who ever lived on earth} & \example{how many people have ever lived on earth} \\
\GEN & \example{price of a wheel for a two thousand five volvo s eighty} & \example{what is the price of a wheel for a car} \\
\GEN & \example{is it good to put a watermelon that is not been peeled in the ice box} & \example{is it good to put watermelon in the ice box} \\
\GEN & \example{how is it another way to defeat hurricane} & \example{how to defeat a hurricane} \\
\midrule
\ROOGEN & \example{in an air fryer at what temperatures do i heat fully cooked bratwurst and for how long}  & \example{how long should bratwurst be cooked} \\
\ROOGEN & \example{in 1901 who won the first nobel prize for physics} & \example{who won the first nobel prize} \\
\ROOGEN & \example{what is the average salary of a taxi driver from san francisco} & \example{how much does a taxi driver earn in california} \\ 
\ROOGEN & \example{in feet how long was the largest shark ever on the record} & \example{what is the longest shark} \\
\ROOGEN & \example{who is the actor's name who plays eleven in stranger things} & \example{who is the actor that plays eleven in stranger things} \\
\ROOGEN & \example{what is a apple or lemon that starts with letter f.} & \example{what is a fruit that start with f} \\
\ROOGEN & \example{does hair grow faster when it is warm or cold outside} & \example{how fast does hair grow} \\
\ROOGEN & \example{can you tell us some facts about patrick mahomes achievements} & \example{what are facts about patrick mahomes} \\
\ROOGEN & \example{if i workout for about ten minutes everyday how many days will it take for me to lose a pound} & \example{how long does it take to lose weight?} \\
\bottomrule
\end{tabular}}
\caption{Synthetic examples of reformulated questions according to the different reformulation operators.}\label{apdx:examples}
\end{table*}

\end{document}